# SituationCO v1.2's Terms, Properties, Relationships and Axioms -- A Core Ontology for Particular and Generic Situations

**Luis Olsina, Guido Tebes,** and **Pablo Becker**
GIDIS_Web, Facultad de Ingeniería, UNLPam, General Pico, LP, Argentina
[olsinal, guido_tebes, beckerp]@ing.unlpam.edu.ar

**Abstract.** The current preprint is an update to SituationCO v1.1 (Situation Core Ontology), which represents its new version 1.2. It specifies and defines all the terms, properties, relationships and axioms of SituationCO v1.2, being an ontology for particular and generic Situations placed at the core level in the context of a four-layered ontological architecture called FCD-OntoArch (*Foundational, Core, and Domain Ontological Architecture for Sciences*). This is a four-layered ontological architecture, which considers Foundational, Core, Domain and Instance levels. In turn, the domain level is split down in two sub-levels, namely: Top-domain and Low-domain ontological levels. So in fact, we can consider it to be a five-tier architecture. Ontologies at the same level can be related to each other, except for the foundational level where only ThingFO (*Thing Foundational Ontology*) is found. In addition, ontologies' terms and relationships at lower levels can be semantically enriched by ontologies' terms and relationships from the higher levels. Note that both ThingFO and ontologies at the core level such as SituationCO, ProcessCO, among others, are domain independent. SituationCO's terms and relationships are specialized primarily from ThingFO. It also completely reuses terms primarily from ProcessCO, ProjectCO and GoalCO ontologies. Stereotypes are the used mechanism for enriching SituationCO terms. Note that in the end of this document, we address the SituationCO vs. ThingFO non-taxonomic relationship verification matrix.

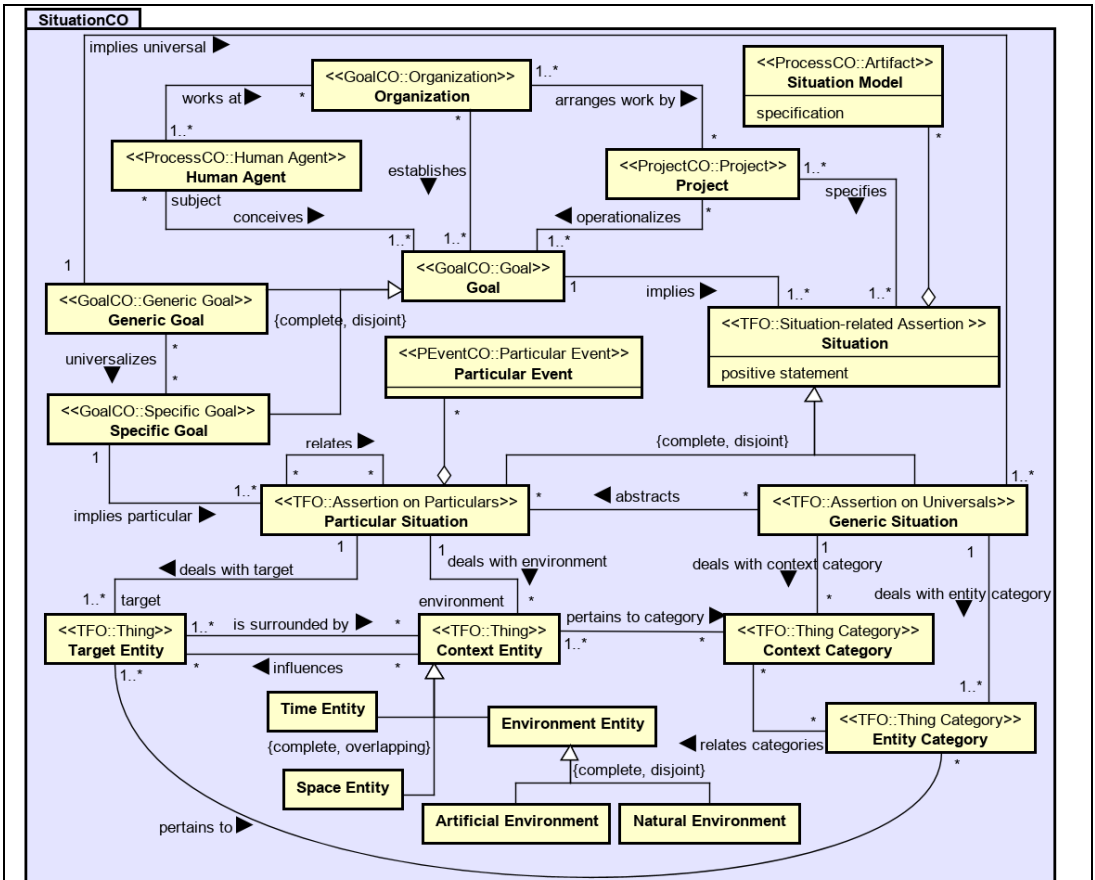

**Figure 1.** SituationCO v1.2: A Core Ontology for Situations, which is placed at the core level of FCD-OntoArch in Fig. 2. Note this is a revised version of SituationCO v1.1 [6]. Annotations of changes/updates from the previous version (v1.1) to the current one (v1.2) can be found in Appendix A.

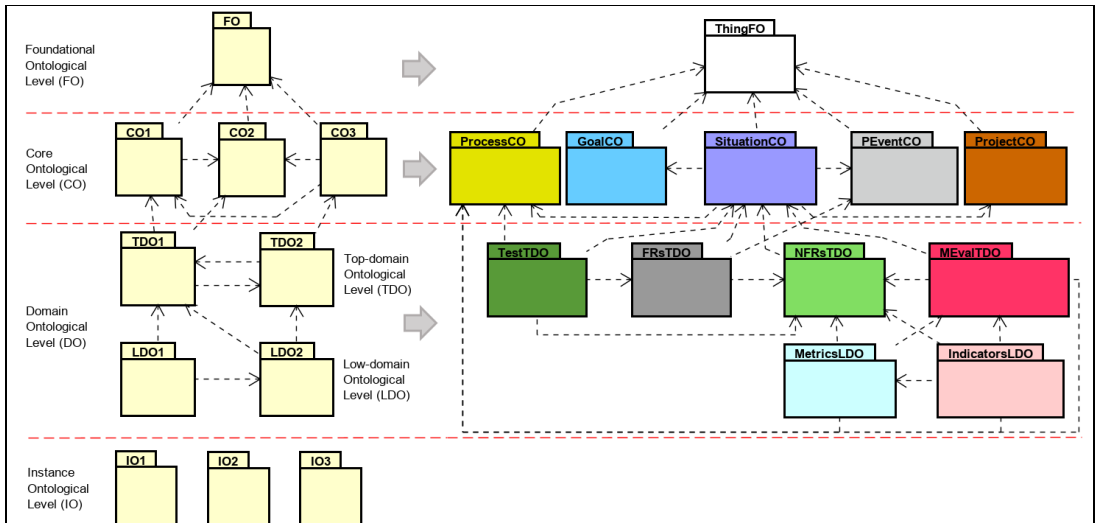

**Figure 2.** Allocating the SituationCO component or module in the context of the five-layered ontological architecture so-called FCD-OntoArch (*Foundational, Core, and Domain Ontological Architecture for Sciences*) [5].

| Situation Component – SituationCO v1.2's Terms ||
|---|---|
| **Term** | **Definition** |
| **Human Agent** | It is an agent, in fact, a human work resource assigned to a work entity, which performs a task in fulfillment of a role. <br><br> Note 1: In contrast, an automated agent is also a work resource that is not a human being, but is made up of hardware and/or software. <br><br> Note 2: A Human Agent embodied by a person –the subject- is the one who conceives Goals. <br><br> Note 3: Human Agent is a term reused completely from ProcessCO [2]. |
| **Organization** | It is a Thing (a particular) comprising people that is structured and managed to establish and pursue organizational Goals and is affected by and affects to its environment or context. |
| **Project** | It is a Thing (a particular) representing a temporary and goal-oriented endeavor with definite start and finish dates, which considers a managed set of interrelated activities, tasks and resources aimed at producing and modifying unique Work Products (i.e., artifacts, services or results) for satisfying a given requester need (Goal). |
| **Goal** <br> (synonym: **Organizational Goal**) | It is an Intention-related Assertion, that is, the statement of the aim to be achieved by the Organization which considers the propositional content of a Goal's purpose in a given Situation and time frame. |
| **Generic Goal** | It is a Goal related to none or more Generic Situations. <br><br> Note: See Note 1 for the Generic Situation term. A synonym of generic is non-specific. |
| **Specific Goal** <br> (synonym: **Objective**) | It is a Goal related to none or more Particular Situations. <br><br> Note: A synonym of specific is non-generic. For example, Specific Goals' purposes are "understand", "monitor", "improve", "find vulnerabilities", "develop", etc., something in particular circumstances. |
| **Situation** (synonym: **Circumstances**, **Phenomenon, State of Affairs**) | It is a Situation-related Assertion that explicitly states and specifies the combination of circumstances, episodes and relationships/events embracing particular entities and their surroundings, or categories of entities and their related generic context, which is of interest and relevant to be represented by a Human Agent/Organization with an established Goal. |
| | Note 1: "Particular entities and their surroundings" have semantic of Things, whereas "categories of entities and their related generic context" have semantic of Thing Categories, both terms coming from the ThingFO ontology [4, 5]. |
| | Note 2: Organizations arrange work through Projects, which define Situations. In turn, an Organization establishes Goals, or more specifically a Human Agent conceives Goals implying Situations. |
| | Note 3: According to Molina *et al.* [3] "*By situation we understand the summing of the entity of interest, the task and intention towards that entity and the interaction between this and other [surrounding] entities*". |
| | Note 4: A Situation can be represented statically or dynamically depending on the intention of the Human Agent/Organization. |
| **Situation Model** | It represents an Artifact that specifies and models Particular or Generic Situations of a given particular or abstract world, respectively. |
| | Note 1: A Situation model can be seen as an explicit representation of a (mental) micro-world of a particular or generic world, which is conceived by a Human Agent with a certain Goal. |
| | Note 2: According to Zwaan "*Situation models are models of events. Events always occur at a certain time and place. In addition, events typically involve participants (agents and patients) and objects. Furthermore, events often entertain causal relations with other events or are part of a goal plan structure. Thus, time, place, participants, objects, causes and effects, and goals and plans are components of situations, with time and place being obligatory.*" [10]. Note that this author quote that the first introduced proposal of the Situation Model in the cognitive psychology area was made by van Dijk & Kintsch [9] in the early 1980s. |
| | Note 3: Situations can be modeled by means of informal, semiformal or formal specification languages. Situation Models can be implemented in Systems for verification and validation purposes. |
| | Note 4: A Situation Model has semantic of Artifact, which is a term coming from ProcessCO [2]. |
| **Particular Situation** | It is a Situation-related Assertion on Particulars that explicitly states and specifies the combination of particular circumstances, episodes and relationships/events embracing Target Entities and their surrounding Context Entities, which is of interest and relevant to be represented by a Human Agent/Organization with an established Specific Goal. |
| | Note 1: See Axioms A1, A2 and A3. |
| | Note 2: Organizations arrange work through Projects. In turn, a Project operationalizes Specific Goals by defining Particular Situations. In other words, an Organization establishes Specific |
| | Goals, or more specifically, a Human Agent working at it conceives Specific Goals implying Particular Situations. |
| **Particular Event** | It is an Assertion on Particulars and, at the same time, an Action-related Assertion that explicitly states and specifies the occurrence of an entity (Thing) action. It is related to the interaction and happening of entities since acting behaviors cause any Particular Events that might occur |
| | Note: Given the nature of the Particular Situation concept, dynamic representations are more complex but richer than static ones, usually including the term Particular Event (see the PEventCO ontology). |
| **Generic Situation** | It is a Situation-related Assertion on Universals that explicitly states and specifies the combination of generic circumstances, episodes and relationships embracing Entity Categories and their related generic Context Categories, which is of interest and relevant to be represented by a Human Agent/Organization with an established Generic Goal. |
| | Note 1: The adjective "generic" is used to describe some concept that refers or relates to a whole class of similar concrete and specific objects, circumstances, features, intentions, etc. It refers to the commonality of a set of something specific. A synonym of generic is non-specific, that is, not specific to any particular Thing, Situation, etc. |
| | Note 2: Organizations arrange work through Projects. In turn, a Project operationalizes Generic Goals by defining Generic Situations. In other words, an Organization establishes Generic Goals, or more specifically, a Human Agent working at it conceives Generic Goals implying Generic Situations. |
| **Target Entity** | It represents a particular or concrete, tangible or intangible object, which is the main focus of interest –the target- for a given Particular Situation. |
| | Note 1: Target Entity has semantic of Thing –term coming from the ThingFO ontology [5, 7]. Therefore, a Target Entity is not a particular object without its properties and powers. |
| | Note 2: Target Entity can be specialized in an Evaluable Entity (term defined in the NFRsTDO component), a Developable Entity (term defined in the FRsTDO component), a Testable Entity (term defined in the TestTDO component), or an Observable Entity. |
| | Note 3: Examples of Evaluable Entities are a concrete Work Process, Product (e.g. model, source code, document), System, Resource, among others. Examples of Developable Entities are a concrete Work Product (e.g. model, source code, document, device, building), System, among others. Examples of Observable Entities can be any object of a particular world that can be investigated. Thus, a case study is an empirical research method which entails Observable Entities. |
| | Note 4: Notice that, for example, a concrete System can be an Evaluable/Developable/Testable/Observable entity, but the instantiated kind of Target Entity depends on the Particular Situation. |
| **Context Entity** | It represents a particular or concrete, tangible or intangible object that surrounds and influences one or more Target Entities, which |
| | depending on a Specific Goal conceived/established by a Human Agent/Organization may be of interest and relevant for a Particular Situation. |
| | Note: Context Entity has semantic of Thing –term coming from the ThingFO ontology [5, 7]. Therefore, a Context Entity is not a particular object without its properties and powers. |
| **Environment Entity** (synonym: **Environment Context Entity**) | It is a Context Entity that represents natural or unnatural objects. |
| **Artificial Environment** (synonym: **Unnatural-Environment Context Entity**) | It is an Environment Context Entity that encompasses particular unnatural objects, that is, those developed by human beings. |
| | Note: Examples of Artificial-Environment Context Entities can be any social, legal, or computerized object, among many others. |
| **Natural Environment** (synonym: **Natural-Environment Context Entity**) | It is an Environment Context Entity that encompasses particular natural objects, that is, those not developed by human beings. |
| | Note: Examples of Natural-Environment Context Entities can be any object such as mineral, air, water, organic, among others. |
| **Space Entity** (synonym: **Place, Location Context Entity**) | It is a Context Entity that represents the limited or unlimited dimensional region or expanse in which all tangible or intangible particulars in a Situation are located and where all interactions and events occur. |
| | Note: The definition above encompasses both physical and virtual spatial objects. |
| **Time Entity** (synonym: **Time Context Entity**) | It is (usually) a Context Entity that represents an intangible particular, a non-spatial continuum in which events in a Situation occur in an apparently irreversible temporal succession from the past to the present and the future. |
| | Note: The phrase above "is usually a Context Entity" reflects the idea that Time may also be the main entity of interest to be investigated, or Time may also represent a work Resource Entity, which is assigned to a Work Entity. |
| **Context Category** (synonym: **Context Entity Category**) | It is a Thing Category to which concrete Context Entities belong to. |
| **Entity Category** | It is a Thing Category to which concrete Target Entities belong to. |
| | Note 1: Universals may be classified in Evaluable Entity Category, Developable Entity Category, Observable Entity Category, among others, depending on the intention of the Human Agent/Organization. |
| | Note 2: Furthermore, common names of Entity Categories of interest documented in diverse scientific disciplines are Organization Category, Project Category, Resource Category, Process Category, Product Category, Service Category, System Category, among others. |

*Amount of Own or Extended Terms: 12 -- Amount of Completely Reused Terms: 8*

| Situation Component – SituationCO v1.2's Attributes or Properties (*) |||
|---|---|---|
| **Term** | **Attribute** | **Definition** |

(*) Note that all Properties of the Terms in SituationCO are reused from the enriching Term (or component), which is indicated in the stereotype. In this way, the Organization term has its defined properties in GoalCO [8], the Project term has its defined properties in ProjectCO [8], the Human Agent has its defined properties in ProcessCO [1, 2], and so forth. Additionally, the Situation term has semantic of Assertion, so it inherits the Assertion properties, which are defined in ThingFO [4, 5]. Likewise, for instance, the Target Entity term, which has semantic of Thing.

| Situation Component – SituationCO v1.2's Non-taxonomic Relationships ||
|---|---|
| **Relationship** | **Definition** |
| **abstracts** | A Generic Situation generalizes Particular Situations. Note: For example, architectural, design, strategy patterns, among others, represent generic, reusable solutions of Particular Situations. |
| **arranges work by** | An Organization organizes its work or effort by means of Projects for the achievement of its established Goals. |
| **conceives** | A Human Agent –the subject- is the one who conceives Goals. Note: A Human Agent may conceive Specific Goals, Generic Goals or both. |
| **deals with context category** | A Generic Situation as Situation-related Assertion deals with none or many Context Categories. |
| **deals with entity category** | A Generic Situation as Situation-related Assertion deals with one or more Entity Categories. |
| **deals with environment** | A Particular Situation as Situation-related Assertion deals with none or many concrete Context Entities. |
| **deals with target** | A Particular Situation as Situation-related Assertion deals with one or more concrete Target Entities. |
| **establishes** | An Organization establishes and pursues Goals as part of its mission, and in alignment with its vision. |
| **implies** | A Goal implies one or more Situations. |
| **implies particulars** | A Specific Goal implies one or more Particular Situations. |
| **implies universals** | A Generic Goal implies one or more Generic Situations. |
| **influences** | A Context Entity influences none or more Target Entities in a Particular Situation. |
| **is surrounded by** | A Target Entity is surrounded by none or more Context |
| | Entities in a Particular Situation. |
| **operationalizes** | A Project operationalizes one or more Goals. <br> Note: A Project, depending on the defined Situation, may operationalize Specific Goals, Generic Goals or both. |
| **pertains to** | A Target Entity may belong to Entity Categories. <br> Note: In other words, an Entity Category predicates about a set of Particulars (Target Entities) or instances. |
| **pertains to category** | A Context Entity may belong to Context Categories. <br> Note: In other words, a Context Category predicates about a set of Particulars (Context Entities) or instances. |
| **relates** | A Particular Situation may be related to another Particular Situations. |
| **relates categories** | An Entity Category may be related to Context Categories in a Generic Situation. |
| **specifies** | A Project defines one or several Situations. <br> Note: A Project may define Particular Situations, Generic Situations or both. |
| **universalizes** | A Generic Goal generalizes none or more Particular Goals. |
| **works at** | A Human Agent usually works in an Organization. |

*Amount of non-taxonomic relationships: 21*

## Situation Component – SituationCO v1.2's Axioms

**A1 description**: *For any Particular Situation that deals with Target Entities, then these Target Entities cannot be a kind of Context Entity.*

**A1 specification**:

$$\forall ps, \forall thing: [ParticularSituation(ps) \land TargetEntity(thing) \\ \land dealsWithTarget(ps, thing) \\ \rightarrow \neg dealsWithEnvironment(ps, thing)]$$

**A2 description**: *For any Particular Situation that deals with Target Entities and Context Entities, then the Target Entities are surrounded by Context Entities.*

**A2 specification**:

$$\forall ps, \forall te, \forall ce: [ParticularSituation(ps) \land TargetEntity(te) \\ \land dealsWithTarget(ps, te) \land ContextEntity(ce) \\ \land dealsWithEnvironment(ps, ce) \rightarrow isSurroundedBy(te, ce)]$$

**A3 description**: *For any Context Entity that influences a Target Entity, then exists a Particular Situation that deals with these Target Entity and Context Entity.*

**A3 specification**:

$$\forall te, \forall ce: [ContextEntity(ce) \land TargetEntity(te) \land influences(ce, te)$$
$$\rightarrow \exists ps: ParticularSituation(ps) \land dealsWithTarget(ps, te)$$
$$\land dealsWithEnvironment(ps, ce)]$$

## SituationCO v1.2 vs. ThingFO v1.2 Non-Taxonomic Relationship Verification Matrix

| \multicolumn{4}{c|}{SituationCO's Non-taxonomic Relationships} | \multicolumn{4}{c}{ThingFO's Non-taxonomic Relationships} |
|---|---|---|---|---|---|---|---|
| card | Term 1 | relationship name | card | Term 2 | card | Term 1 | relationship name | card | Term 2 |

| card | Term 1 | relationship name | card | Term 2 | card | Term 1 | relationship name | card | Term 2 |
|---|---|---|---|---|---|---|---|---|---|
| 1 | Generic Goal | implies universal | 1..* | Generic Situation | * | Assertion on Universals | relates with | * | Assertion on Universals |
| 1..* | Human Agent | works at | * | Organization | 1..* | Thing | relates with | 1..* | Thing |
| * | Human Agent | conceives | 1..* | Goal | * | Thing | defines | * | Assertion |
| * | Organization | establishes | 1..* | Goal | * | Thing | defines | * | Assertion |
| 1..* | Organization | arranges work by | * | Project | 1..* | Thing | relates with | 1..* | Thing |
| * | Project | operationalizes | 1..* | Goal | * | Thing | defines | * | Assertion |
| 1..* | Project | specifies | 1..* | Situation | * | Thing | defines | * | Assertion |
| 1 | Goal | implies | 1..* | Situation | * | Assertion | relates with | * | Assertion |
| 1 | Specific Goal | implies particulars | 1..* | Particular Situation | * | Assertion on Particulars | relates with | * | Assertion on Particulars |
| * | Generic Goal | universalizes | * | Specific Goal | * | Assertion on Universals | generalizes | * | Assertion on Particulars |
| * | Particular Situation | relates | * | Particular Situation | * | Assertion on Particulars | relates with | * | Assertion on Particulars |
| 1 | Particular Situation | deals with target | 1..* | Target Entity | 1..* | Assertion on Particulars | deals with particulars | 1..* | Thing |
| 1 | Particular Situation | deals with environment | * | Context Entity | 1..* | Assertion on Particulars | deals with particulars | 1..* | Thing |
| * | Generic Situation | abstracts | * | Particular Situation | * | Assertion on Universals | generalizes | * | Assertion on Particulars |
| 1 | Generic Situation | deals with context category | * | Context Category | 1..* | Assertion on Universals | deals with universals | 1..* | Thing Category |
| 1 | Generic Situation | deals with entity category | 1..* | Entity Category | 1..* | Assertion on Universals | deals with universals | 1..* | Thing Category |
| * | Entity Category | relates categories | * | Context Category | * | Thing Category | relates with | * | Thing Category |
| 1..* | Context Entity | pertains to category | * | Context Category | 1..* | Thing | belongs to | * | Thing Category |
| 1..* | Target Entity | is surrounded by | * | Context Entity | 1..* | Thing | relates with | 1..* | Thing |
| * | Context Entity | influences | * | Target Entity | 1..* | (Power of) Thing | interacts with other | 1..* | Thing |
| 1..* | Target Entity | pertains to | * | Entity Category | 1..* | Thing | belongs to | * | Thing Category |

**Acknowledgments.** We warmly thank Maria Julia Blass and Silvio Gonnet (both CONICET researcher and professor at the National University of Technology, Santa Fe, Argentina) for the validation and feedback provided, which allowed us to improve SituationCO from v1.1 to v1.2.

# Appendix A: Updates from SituationCO v1.1 to SituationCO v1.2

Note that the previous version of the SituationCO ontology (i.e., v1.1) is in [6].

- The addition of three axioms, currently labeled A1, A2 and A3, which are specified in first-order logic. SituationCO v1.1 had no specified axiom.
- The addition of new constraints expressed in the diagram with the labels {`complete, disjoint`} and {`complete, overlapping`}, as depicted in Fig. 1.
- The addition of the SituationCO v1.2 vs. ThingFO v1.2 non-taxonomic relationship verification matrix. As a consequence, many non-taxonomic relationships were renamed in order to reflect the reuse by refinement of the corresponding ThingFO relationships. Also, a couple of cardinalities were changed.
- The addition of the term Particular Event, which is reused completely from the PEventCO component.
- The addition of the non-taxonomic relationship 'universalizes' between Specific Goal and Generic Goal.